%% file: main.tex
\documentclass[lettersize,journal]{IEEEtran}
\usepackage{amsmath,amsfonts}
\usepackage{algorithmic}
\usepackage{algorithm}
\usepackage{array}
\usepackage{textcomp}
\usepackage{stfloats}
\usepackage{url}
\usepackage{verbatim}
\usepackage{graphicx}
\usepackage{cite}
\usepackage{amssymb}
\usepackage{booktabs}
\usepackage{algorithmic}
\usepackage[switch]{lineno}
\usepackage{multirow}
\usepackage{colortbl}
\usepackage{enumitem}
\usepackage{subcaption}
\definecolor{mycolor}{rgb}{0,0,0}
\begin{document}

\title{Enhancing Ambiguous Dynamic Facial Expression Recognition with Soft Label-based Data Augmentation}




\author{Ryosuke Kawamura, Hideaki Hayashi, Shunsuke Otake, Noriko Takemura, Hajime Nagahara
\thanks{Ryosuke Kawamura is with the Converging Lab, Fujitsu Research of America, Inc., 115 Atwood, Pittsburgh, PA, 15213, U.S. Email: rkawamura@fujitsu.com}
\thanks{Hideaki Hayashi and Hajime Nagahara are with the D3 Center, the University of Osaka, 2-8 Yamadaoka, Suita, Osaka 565-0871, Email: \{hayashi, nagahara\}@ids.osaka-u.ac.jp}
\thanks{Noriko Takemura is with the Graduate School of Computer Science and Systems Engineering, Kyushu Institute of Technology, 680-4 Kawazu, Iizuka, Fukuoka 820-8502, Japan, Email:takemura@ai.kyutech.ac.jp}}

\markboth{Journal of \LaTeX\ Class Files,~Vol.~14, No.~8, August~2021}%
{Shell \MakeLowercase{\textit{et al.}}: A Sample Article Using IEEEtran.cls for IEEE Journals}


\maketitle

\begin{abstract}
Dynamic facial expression recognition (DFER) is a task that estimates emotions from facial expression video sequences. For practical applications, accurately recognizing ambiguous facial expressions---frequently encountered in in-the-wild data---is essential. In this study, we propose MIDAS, a data augmentation method designed to enhance DFER performance for ambiguous facial expression data using soft labels representing probabilities of multiple emotion classes. MIDAS augments training data by convexly combining pairs of video frames and their corresponding emotion class labels. This approach extends mixup to soft-labeled video data, offering a simple yet highly effective method for handling ambiguity in DFER. To evaluate MIDAS, we conducted experiments on both the DFEW dataset and FERV39k-Plus, a newly constructed dataset that assigns soft labels to an existing DFER dataset. The results demonstrate that models trained with MIDAS-augmented data achieve superior performance compared to the state-of-the-art method trained on the original dataset.

\end{abstract}


\section{Introduction}
\input{src/1_Introduction.tex}

\section{Related Work}
\input{src/2_RelatedWork.tex}

\section{MIDAS: Mixing Ambiguous Data with Soft Labels}
\input{src/3_Method.tex}

\section{Experiments}
\input{src/4_Experiments.tex}

\section{Analysis and Ablation Studies}
\input{src/5_AblationStudy.tex}

\section{Conclusion}
\input{src/6_Conclusion.tex}

\bibliographystyle{IEEEtran}
\bibliography{refs}
\newpage

 




\vfill

\end{document}

%% file: src/1_Introduction.tex
\label{sec:intro}
\IEEEPARstart{F}{acial} expressions are essential for human communication, and facial expression recognition has broad applications in fields such as human-computer interaction, driver monitoring, and intelligent tutoring systems for education. Understanding emotions accurately requires considering the temporal dynamics of facial expressions, as they are driven by facial muscle movements, as shown in previous studies~\cite{liu2014learning,yang2008facial,yang2009boosting}. Based on this, our research focuses on dynamic facial expression recognition (DFER), which involves identifying an emotion class from a video clip.

\begin{figure}[t]
  \centering
  \includegraphics[width=0.95\linewidth]{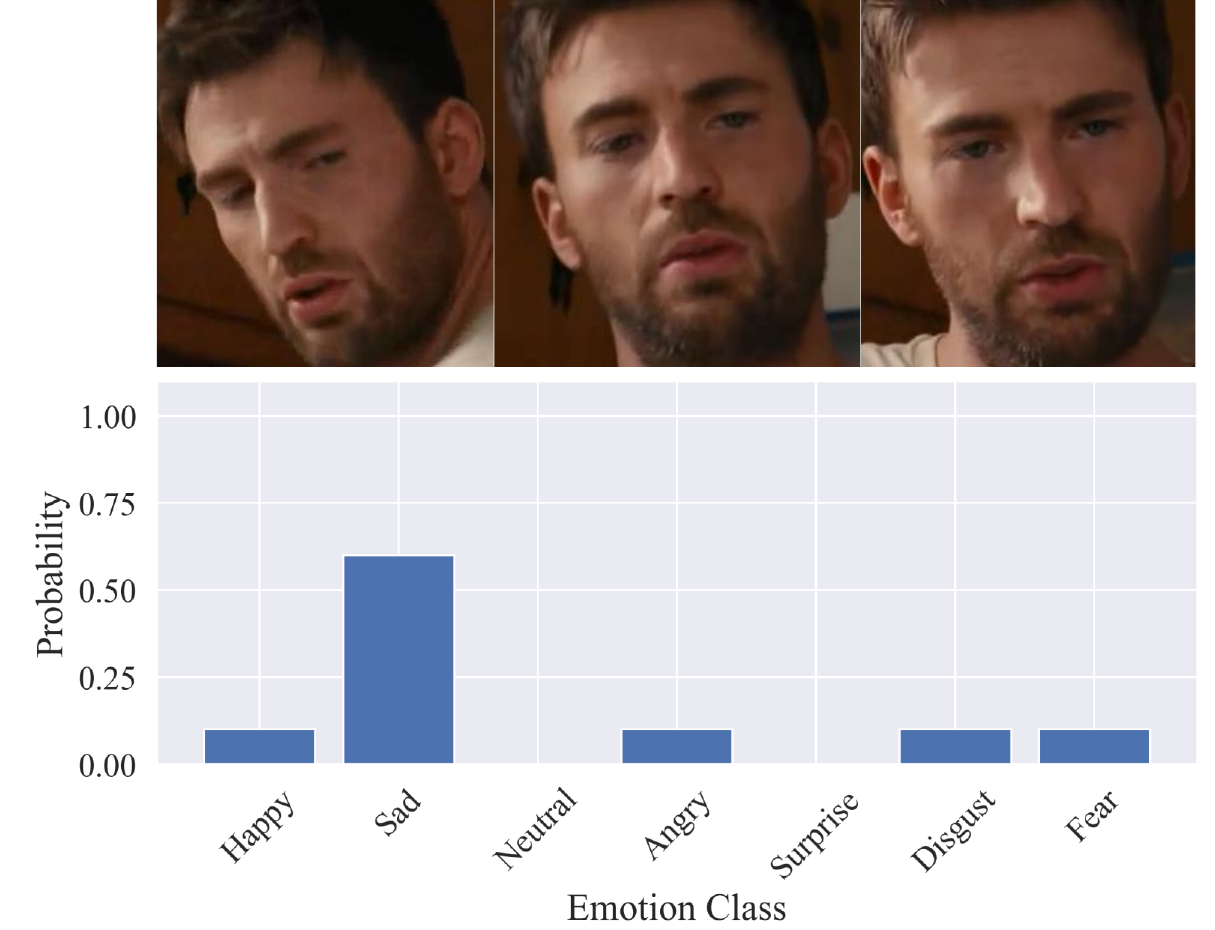}
  \vspace{0pt}
  \caption{Example of an ambiguous facial expression. The images were taken from the DFEW dataset~\protect\cite{jiang2020dfew}.
  The bar chart in the bottom row represents the soft-labeled annotation, constructed based on the proportion of votes from ten annotators. The annotations are distributed across five emotion classes.
}
  \vspace{0pt}
  \label{fig:ex}
\end{figure}

Although deep learning-based techniques have shown remarkable performance in DFER on lab-controlled data, recognizing dynamic facial expressions in in-the-wild settings remains challenging. This difficulty arises from the presence of ambiguous facial expressions that cannot be clearly categorized in a single emotion class.
Several factors contribute to this ambiguity, with the coexistence of multiple emotions being one of the major causes. 
Since emotions are not mutually exclusive but collective, multiple emotions can coexist at different intensities in ambiguous facial expressions captured under natural conditions. 
This differs significantly from lab-controlled data, where subjects are instructed to display specific expressions. Moreover, facial expressions evolve over time, leading to the presence of multiple emotions even within a single video clip.
For example, in Fig.~\ref{fig:ex}, the annotators' evaluations for this video clip were split between ``Happy,'' ``Sad,'' ``Angry,'' ``Disgust,'' and ``Fear'' with different probabilities. Similarly, when witnessing an unjust act, an individual's expression may simultaneously convey anger and disgust. Such variations in emotional perception and facial expressions contribute significantly to the ambiguity in annotation.

Attaching soft labels to training data, instead of hard labels, is an effective way to address the ambiguity in DFER. Hard labels, that is, one-hot encoded class labels, are mostly used in general recognition tasks such as object recognition, where the input sample is clearly categorized into a single class; however, they cannot appropriately represent a combination of multiple emotions with different intensities in ambiguous facial expressions. 
{\color{mycolor}
In contrast, soft labels reflect the possibility of multiple co-occurring emotions.. Rather than assigning a fixed class, they assign probability values to different emotions, reflecting the inherent variability in human perception. To effectively learn the ambiguity in DFER, leveraging soft labels that encode probabilistic distributions of emotions allows for a more comprehensive use of the information provided by annotators. }
One possible method of assigning soft labels is to have multiple annotators evaluate the training data and use the ratio of their votes.
{\color{mycolor} For example, in Fig.~\ref{fig:ex}, 60\% of annotators assigned the label ``Disgust,'' while 20\% selected ``Neutral,'' and 10\% voted for ``Fear'' and ``Angry.''}

One disadvantage of soft labels is that their greater flexibility compared to hard labels makes it challenging to ensure consistency across diverse patterns. 
{\color{mycolor}The vast number of possible combinations of emotion classes and their corresponding probabilities presents a significant challenge in constructing comprehensive training data. Covering all potential patterns is impractical. Generating soft labels requires multiple annotators per sample, which significantly increases labeling costs. As a result, it becomes infeasible to fully capture the entire range of emotional variations in the dataset.} 
Furthermore, the size of the dataset is often limited in DFER due to the difficulty of manual annotation and data collection. To address this problem, effective and well-designed data augmentation is essential to enhance model performance under limited data conditions.


\IEEEpubidadjcol
In this paper, we introduce MIDAS (\underline{mi}xing ambiguous \underline{da}ta with \underline{s}oft labels), a data augmentation method designed for DFER with ambiguous facial expressions. MIDAS extends the mixing strategy to accommodate both soft labels and dynamic facial expressions. The method combines pairs of video frames representing facial expressions with the corresponding soft labels, which indicate the probabilities of emotion classes. The model is then trained on the generated data.

{\color{mycolor}
This paper is an extension of our previous conference publication~\cite{kawamura2024midas}. This study introduces two key advancements over our previous work. First, to assess generalizability across datasets, we construct a new soft-labeled dataset by incorporating additional annotations into FERV39k~\cite{wang2022ferv39k}, a large-scale hard-label dataset for the DFER task. Second, to evaluate the generalizability across model architectures, we perform experiments using multiple network architectures.
}
Our contributions are summarized as follows, where * denotes new contributions of this paper:
\begin{itemize}
    \item We proposed a data augmentation method for DFER with ambiguous facial expressions called MIDAS. MIDAS convexly combines pairs of video frames representing facial expressions in a manner similar to mixup. A key significant difference from mixup is that MIDAS can be applied when the true hard labels are unavailable, and only soft labels, which consist of probabilities for multiple classes, are given. 
    \item We demonstrated that MIDAS corresponds to minimizing the vicinal risk in cases where the true hard label is unknown with a vicinity distribution using a random ratio and virtual labels that are different from the original mixup.
    \item * {\color{mycolor} We constructed a new dataset by extending FERV39k through the addition of soft labels. Annotations are collected from multiple annotators via crowdsourcing, and the voting results are incorporated as soft labels to enhance the dataset’s representation of ambiguous facial expressions.}
    \item * We demonstrated that the proposed method outperforms existing state-of-the-art approaches through DFER experiments on the DFEW dataset. Additionally, the ablation study revealed that the combination of soft labels and mixing strategy has a synergistic effect although each method is also effective when applied independently. 
\end{itemize}

%% file: src/2_RelatedWork.tex
\subsection{DFER}\label{sec:rw-dfer}
While many in-the-wild datasets for static FER use images sourced from the internet, most DFER datasets, such as CK+~\cite{luceyExtendedCohnKanadeDataset2010} and Oulu-CASIA~\cite{zhao2011facial}, are collected in controlled laboratory settings, where facial expressions are elicited through researcher instructions. 
Recently, there has been an increasing effort to develop large-scale in-the-wild datasets for DFER. AFEW, introduced by Dhall \textit{et al.}~\cite{dhallCollectingLargeRichly2012}, was the first in-the-wild DFER dataset, consisting of short movie clips annotated by two annotators. Jiang \textit{et al.}~\cite{jiang2020dfew} also compiled movie clips for the DFEW dataset, where each video was annotated by ten out of twelve available annotators, providing both single-labeled and seven-dimensional emotion class annotations. 
FERV39k~\cite{wang2022ferv39k} is another large-scale in-the-wild DFER dataset, featuring video clips across 22 fine-grained contexts, including business, daily life, and school. These clips were annotated by 20 crowdworkers and 10 professional researchers. Additionally, Liu \textit{et al.} introduced MAFW~\cite{liu2022mafw}, a multimodal affective database comprising video-audio clips collected in the wild. Each clip is annotated with a compound emotional category and descriptive sentences detailing the subjects' affective behaviors.

Regarding FER in the video, methods based on selecting peak frames or aggregating features from each frame have been proposed by~\cite{zhao2016peak,meng2019frame,knyazev2017convolutional,yu2018deeper,yang2018facial,kumar2020noisy}. Two- or three-dimensional convolutional neural networks (2D-CNN or 3D-CNN) combined with a sequential neural network, such as long short-term memory (LSTM) and gated recurrent unit (GRU), are commonly used in~\cite{kim2017multi,yan2018multi,chen2020stcam,kim2017deep,vielzeuf2017temporal,zhang2020facial,liu2020saanet,aminbeidokhti2019emotion}. 
Several studies have explored the use of Transformer-based modules for Dynamic Facial Expression Recognition (DFER)~\cite{zhao2021former,ma2022spatio}. Zhao \textit{et al.}~\cite{zhao2021former} introduced Former-DFER, a model built on the Transformer architecture. Ma \textit{et al.}~\cite{ma2022spatio} proposed a spatio-temporal Transformer (STT) that processes features extracted by a 2D-CNN. Wang \textit{et al.}~\cite{wang2022dpcnet} developed the Dual Path Multi-Excitation Collaborative Network (DPCNet), which integrates a spatial-frame excitation module for spatial feature extraction and a channel-temporal aggregation module to capture channel and temporal dependencies. 
Her et al.~\cite{her2025fru} proposed the FRU-Adapter, which integrates a frame recalibration unit to enhance the contribution of informative frames while reducing the influence of less relevant ones. 
S2D~\cite{chen2024S2D} employs a landmark-aware image model for facial expression recognition in videos, utilizing Multi-View Complementary Prompters and an Emotion-Anchors-based Self-Distillation Loss to enhance performance. Wang \textit{et al.}~\cite{wang2023rethinking} introduced a novel learning paradigm for DFER based on a Multi-Instance Learning pipeline.
Self-training strategies have also been explored for DFER. For instance, MAE-DFER~\cite{sun2023mae} is a self-supervised method incorporating a local-global interaction Transformer and explicit temporal motion modeling. SVFAP~\cite{sun2024svfap} and FE-Adapter~\cite{gowda2024fe} significantly reduce computational costs while maintaining strong performance.

Some studies have investigated the potential of leveraging multimodal information by focusing on the multimodal nature of facial videos. HiCMAE~\cite{sun2024hicmae} is a self-supervised framework for audio-visual emotion recognition that employs masked data modeling and contrastive learning. MMA-DFER~\cite{chumachenko2024mma} also integrates audio-visual information but does not require large-scale paired multimodal pretraining or pretraining for static facial expression recognition. 
Maria \textit{et al.}~\cite{foteinopoulou2024emoclip} introduce EmoClip, which employs sample-level text descriptions as natural language supervision for zero-shot classification.
FineCLIPER~\cite{chen2024finecliper} introduces a vision-text learning paradigm that enhances recognition performance by incorporating both positive and negative textual descriptions of class labels along with fine-grained descriptions of facial dynamics.

\subsection{Ambiguity in FER}\label{sec:rw-amb}
Facial expressions often convey multiple emotions simultaneously~\cite{zhouEmotionDistributionRecognition2015,dantcheva2017expression}. Ambiguous data are typically treated as noisy or inconsistent, and various methods have been proposed to address this issue, including uncertainty estimation. She \textit{et al.}~\cite{sheDiveAmbiguityLatent2021} introduced an architecture incorporating a latent label distribution module and an uncertainty estimation module to handle ambiguity. Wang \textit{et al.}~\cite{wang2020suppressing} designed an additional module to suppress misleading instances and identify latent truths using estimated confidence levels.

Several studies have explored approaches to mitigate label noise and ambiguity. Landmark-Aware Net~\cite{wu2023net} leverages facial landmarks information to combat label noise. Le \textit{et al.}~\cite{le2023uncertainty} proposed a method that constructs adaptive emotion distributions by incorporating neighborhood information in the valence-arousal space. 
Li \textit{et al.}~\cite{li2023intensity} introduced a global convolution attention block and intensity-aware loss (GCA+IAL), where IAL focuses the network on the most challenging categories.
While their approach addresses uncertainty through attention mechanisms and loss functions, our method tackles ambiguity in DFER using data augmentation with soft labels. Soft labels provide a straightforward solution for handling ambiguous data. Barsoum \textit{et al.}~\cite{zhouEmotionDistributionRecognition2015} demonstrated that soft labels annotated by multiple crowd workers improved static FER performance over hard labels. Additionally, Gan \textit{et al.}~\cite{ganFacialExpressionRecognition2019} proposed a framework to generate pseudo-soft labels for static FER. However, to the best of our knowledge, no previous research has explored the use of a mixing strategy with soft labels for DFER.

\subsection{Mixing strategy}\label{sec:rw-mix}
The application of mixing strategies in data augmentation has been extensively studied~\cite{cao2024survey}. Zhang \textit{et al.}~\cite{zhang2018mixup} introduced \textit{mixup}, a method that synthesizes additional training samples by convexly combining random image pairs and their corresponding labels. Based on the vicinal risk minimization principle~\cite{chapelle2000vicinal}, mixup approximates the true data distribution by leveraging the vicinity of each training sample, thereby improving generalization. Thulasidasan \textit{et al.}~\cite{thulasidasan2019mixup} further demonstrated that mixup enhances confidence calibration during training.

Inspired by mixup, various mixing-based augmentation techniques have been proposed. AugMix~\cite{hendrycks2019augmix} and AugMax~\cite{wang2021augmax} combine randomly selected transformations, while PixMix~\cite{hendrycks2022pixmix} and IPMix~\cite{huang2023ipmix} incorporate external datasets for mixing. DiffuseMix~\cite{islam2024diffusemix} generates augmented samples using diffusion-based methods. CutMix~\cite{yun2019cutmix} employs regional crop-and-paste techniques and has been widely adopted~\cite{devries2017improved,lee2020smoothmix,han2022you,qin2024sumix,chen2024cutfreq,baek2021gridmix,rame2021mixmo}. Several methods utilize saliency information, including SaliencyMix~\cite{uddin2020saliencymix}, PuzzleMix~\cite{kim2020puzzle}, ResizeMix~\cite{qin2020resizemix}, and Co-Mixup~\cite{kim2021co}. Additionally, activation or attention maps guide mixing in approaches such as Attentive-CutMix~\cite{walawalkar2020attentive}, TransMix~\cite{chen2022transmix}, TokenMix~\cite{liu2022tokenmix}, and GuidedMixup~\cite{kang2023guidedmixup}. Some methods apply mixing in feature space, including Manifold-Mix~\cite{verma2019manifold}, PatchUp~\cite{faramarzi2020patchup}, and Catch-Up Mix~\cite{kang2024catch}. MixGen~\cite{hao2023mixgen} improves vision-language representation learning by applying multi-modal data augmentation through image interpolation and text concatenation. Unlike our approach, which focuses on label-based classification, MixGen is designed for tasks such as visual grounding and reasoning.

While most studies focus on image-based mixing, few methods extend these strategies to video data. Sahoo \textit{et al.}~\cite{sahoo2021contrast} introduced background mixing for contrastive learning in action recognition, though their approach did not generate data belonging to a different class. Additionally, existing mixing-based studies rely on hard labels, as class boundaries are clearly defined in tasks such as object and action recognition. To the best of our knowledge, no prior research has explored a mixing strategy incorporating soft labels for DFER.

%% file: src/3_Method.tex
\begin{figure*}[t]
  \centering
  \includegraphics[width=0.8\linewidth]{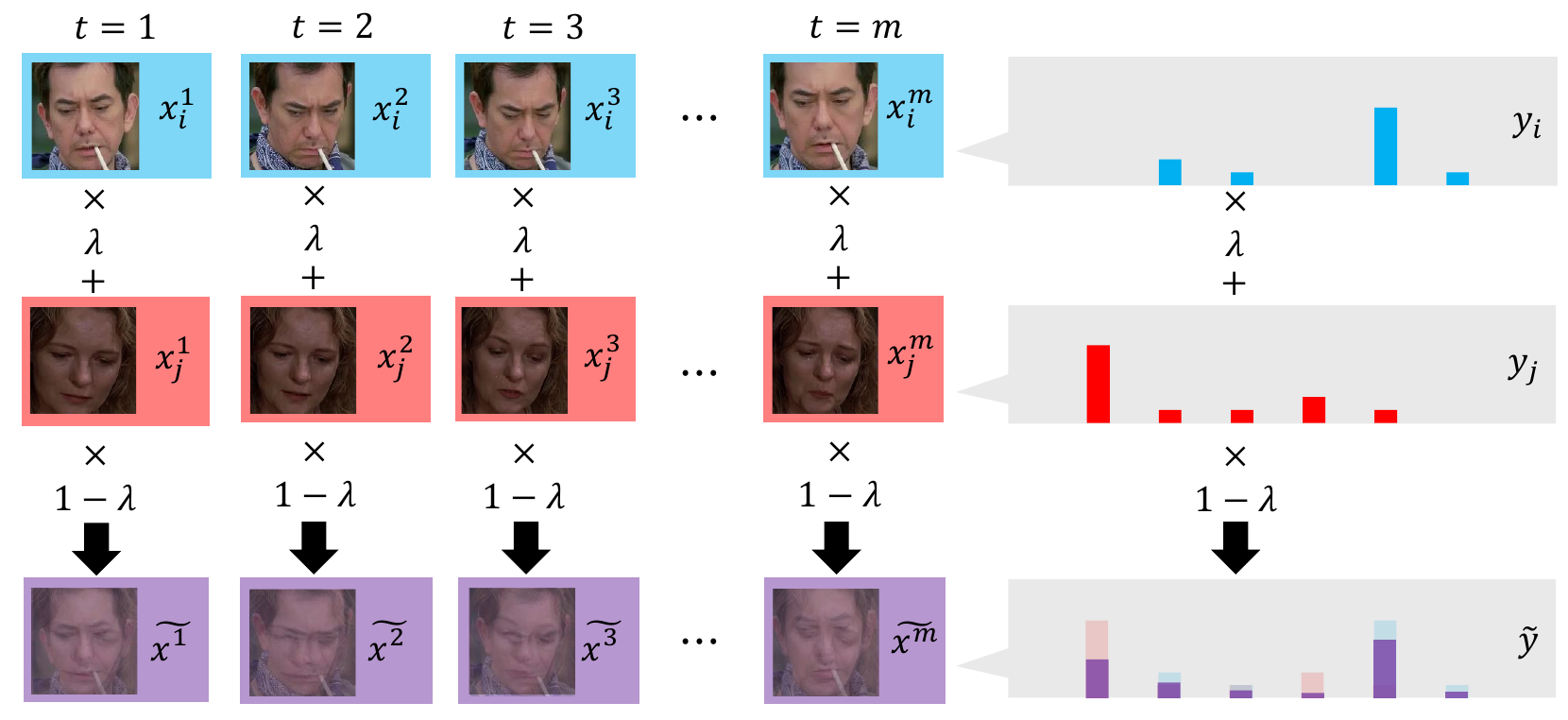}
  \vspace{0pt}
  \caption{Outline of the data mixing procedure in MIDAS. In MIDAS, training data are augmented by convexly combining pairs of video frames along with their corresponding emotion class labels. The mixing coefficient $\lambda$ is randomly sampled from a beta distribution. A key feature of this approach is the use of soft labels, which represent class probabilities, instead of hard labels.}
  \vspace{0pt}
  \label{fig:mixup}
\end{figure*}

MIDAS generates data similarly to mixup by convexly combining given training data and labels using a randomly generated mixing coefficient. It differs from mixup in two key aspects: (i) the input data are video clips, and (ii) soft labels, which represent class probabilities, are used instead of single ground-truth class labels encoded in one-hot format, i.e., hard labels. The soft labels are assumed to be given based on the average votes from multiple annotators. This is due to the difficulty in recognizing ambiguous facial expressions, even for humans, where each annotator's judgment may not always be correct.

The crucial key aspect is that the true hard label is unknown, and the proposed method aims to minimize the vicinal risk under this condition. The data mixing procedure and how it minimizes vicinal risk are detailed below.

\subsection{Data mixing}\label{sec:datamixing}
Let $X_{i}=\left(x^{(1)}_i,\ldots,x^{(T)}_i\right)$ represent the $i$-th video clip in the training dataset, where $T$ is a length of the clip. Here, $x^{(t)}_i \in \mathbb{R}^{H\times W \times 3}$ is an image at the $t$-th frame in the video, with height $H$ and width $W$. 
Additionally, let $y_i \in \mathbb{R}^C$ denote the soft-labeled ground truth for the $i$-th video clip, where its elements represent the probabilities for $C$ emotion classes.
MIDAS generates virtual samples by combining each frame of two video clips randomly selected from the training dataset, as illustrated in Fig.~\ref{fig:mixup}. 
The generated video clip $\Tilde{X}$ and label $\Tilde{y}$ are defined as
\begin{flalign}
\Tilde{X}&=\left(\Tilde{x}^{(1)},\ldots,\Tilde{x}^{(T)}\right),\\
\Tilde{x}^{(t)}&=\lambda x^{(t)}_i + (1-\lambda)x^{(t)}_j,\\
\Tilde{y}&=\lambda y_i + (1-\lambda)y_j,
\end{flalign}
where $\lambda \in [0,1]\sim \beta (\alpha,\alpha)$ represents a random ratio sampled from a beta distribution with parameter $\alpha$. It is important to note that MIDAS does not allow the same video to be selected. 
After that, to account for annotation noise, such as misjudgment~\cite{uma2020case}, a combined soft label $\Tilde{y}$ is normalized using a softmax operation. By applying our method to the facial expression data, it becomes possible to generate data containing multiple emotion classes with varying intensities and temporal changes.

\subsection{Vicinal risk minimization}
The data augmentation performed by MIDAS is justified through the viewpoint of vicinal risk minimization~\cite{chapelle2000vicinal}, similar to the approach used in mixup~\cite{zhang2018mixup}. The key distinction from mixup is that the true hard label for each training sample is unknown. Instead, MIDAS uses a soft label that includes the variation of the annotators' evaluation. As a result, MIDAS can be explained as calculating an empirical risk using a distribution that differs from that of mixup.

In supervised learning, a joint probability distribution $P(x,y)$ is assumed over input and output variables. The objective is to minimize the expectation of a given loss function $\ell$.
\begin{equation}
\label{eq:riskMimization}
    R(f) = \int\ell(f(x),y)\mathrm{d}P(x,y),
\end{equation}
where $f$ denotes a classifier to be trained. However, eq.~(\ref{eq:riskMimization}) cannot be computed directly because the joint distribution $P(x,y)$ is unknown. In general, the empirical risk is minimized instead based on a training dataset $\{\left(x_i,y_i\right)\}^M_{i=1} \sim P(x,y)$.
\begin{equation}
\label{eq:empiricalRisk}
    R_\mathrm{emp}(f) = \frac{1}{M}\sum^M_{i=1}\ell(f(x_i),y_i)
\end{equation}
The empirical risk is obtained by taking an expectation of the loss function over an empirical distribution $P_\mathrm{emp}(x,y) = \frac{1}{M}\sum^M_{i=1}\delta(x=x_i,y=y_i)$, where $\delta$ is a Dirac measure. 

The various choices exist for approximating the true distribution, and selecting a different distribution leads to alternative methods of risk minimization. One such approach is the empirical vicinal risk, which is based on the vicinity distribution~\cite{chapelle2000vicinal}. In~\cite{zhang2018mixup}, it was demonstrated that mixup training minimizes the empirical vicinal risk:
\begin{equation}
    R_\mathrm{mixup}(f) = \frac{1}{M}\sum^M_{i=1}\ell(f(\tilde{x}_i),\tilde{y}_i),
\end{equation}
where $\{(\tilde{x}_i,\tilde{y}_i)\}^M_{i=1}$ denotes a set of virtual feature-target pairs generated from the vicinity distribution defined as
\small
\begin{flalign}
\label{eq:VicinityDistribution}
    &P_{\mathrm{mixup}}(\tilde{x}_i,\tilde{y}_i \mid x_i,y_i) \nonumber \\
    &\!=\! \frac{1}{n}\sum^n_j\underset{\lambda}{\mathbb{E}}[\delta(\tilde{x}_i \!=\! \lambda x_i \!+\! (1-\lambda)x_j, \tilde{y}_i \!=\! \lambda y_i \!+\! (1-\lambda)y_j)].
\end{flalign}
\normalsize

In the context of this study, the hard label $y_i$\footnote{The superscript for the frame number is omitted for simplicity in this subsection.} corresponding to the underlying true emotion is unknown. Instead, a soft label $q_i$ is provided, which is based on the voting average of multiple annotators. In MIDAS, the training data are sampled from the following distribution:
\small
\begin{flalign}
\label{eq:softVicinityDistribution}
    &P_{\mathrm{MIDAS}}(\tilde{x}_i,\tilde{y}_i \mid x_i,q_i) \nonumber \\
    &\!=\! \frac{1}{n}\sum^n_j\underset{\lambda}{\mathbb{E}}[\delta(\tilde{x}_i \!=\! \lambda x_i \!+\! (1-\lambda)x_j, \tilde{y}_i \!=\! \lambda q_i \!+\! (1-\lambda)q_j)].
\end{flalign}
\normalsize

Equation~(\ref{eq:softVicinityDistribution}) can be viewed as a variation of vicinity distribution. Assuming that $S$ annotators assign one-hot labels $v_i^{(s)}$ ($s = 1, \ldots, S$) to each training sample $x_i$, the soft label $q_i$ is computed by the average of the annotators' votes, given by $q_i = \frac{1}{S}\sum^S_{s=1}v_i^{(s)}$. If $l$ out of $S$ votes are correct, $q_i$ is expressed as $q_i = \frac{l}{S}y_i + \frac{1}{S}\sum_{s \in \mathcal{W}}v_i^{(s)}$, where $\mathcal{W}$ represents a set of indices for wrong annotations. Using this expression, Eq.~(\ref{eq:softVicinityDistribution}) can be written as 
\small
\begin{flalign}
\label{eq:newSoftVicinityDistribution}
    &P_{\mathrm{MIDAS}}(\tilde{x}_i,\tilde{y}_i \mid x_i,q_i) \nonumber \\
    &\!=\! \frac{1}{n}\!\sum^n_j\underset{\lambda,\lambda'}{\mathbb{E}}\![\delta(\tilde{x}_i \!\!=\!\! \lambda x_i \!+\! (1\!-\!\lambda)x_j, 
    \tilde{y}_i \!\!=\!\! \lambda' y_i \!+\! (1\!-\!\lambda')y'_j)],
\end{flalign}
\normalsize
where we defined $\lambda' = \frac{\lambda l}{S}$ and $y'_j = \frac{\lambda}{S-\lambda l}\sum_{s \in \mathcal{W}}v_i^{(s)} + \frac{S(1-\lambda)}{S-\lambda l}q_j$. 
MIDAS corresponds to minimizing the vicinal risk in a situation where the true hard label $y_i$ is unknown. This is achieved by defining the vicinity distribution using a random ratio and virtual labels that differ from the original mixup.

%% file: src/4_Experiments.tex
\label{sec:experiments}
The objective of this experiment is to assess the effectiveness of MIDAS for DFER using a deep learning-based automatic DFER model. For this experiment, we first created a new dataset by assigning soft labels to an existing DFER dataset that originally only contained hard labels. Second, to evaluate the performance of MIDAS, we used both a publicly available DFER dataset and our newly created dataset. The results were compared with those of existing DFER methods, including the state-of-the-art approach.

\subsection{Evaluation dataset}
We use two datasets for our experiment.
\subsubsection{DFEW}
The dynamic facial expression in-the-wild (DFEW) dataset consists of 11,967 video clips sourced from movies. The DFEW is unique in that it provides soft labels for each individual video clip. These clips present various challenges in practical scenarios, including extreme illumination, occlusions, and capricious pose changes.  For this dataset, twelve expert annotators were employed, and ten out of twelve annotators were assigned to each video clip. Each annotator was tasked with selecting one out of seven emotion classes (``Happy,'' ``Sad,'' ``Neutral,'' ``Angry,'' ``Surprise,'' ``Disgust,'' and ``Fear''). The resulting votes from the annotators are provided as seven-dimensional emotion distribution labels, which serve as soft labels in this experiment. Additionally, his dataset includes the class with the highest number of votes by the annotators for each clip, which we used as a hard label in comparative experiments.\par

\begin{figure}[t]
  \centering
  \includegraphics[width=0.95\linewidth]{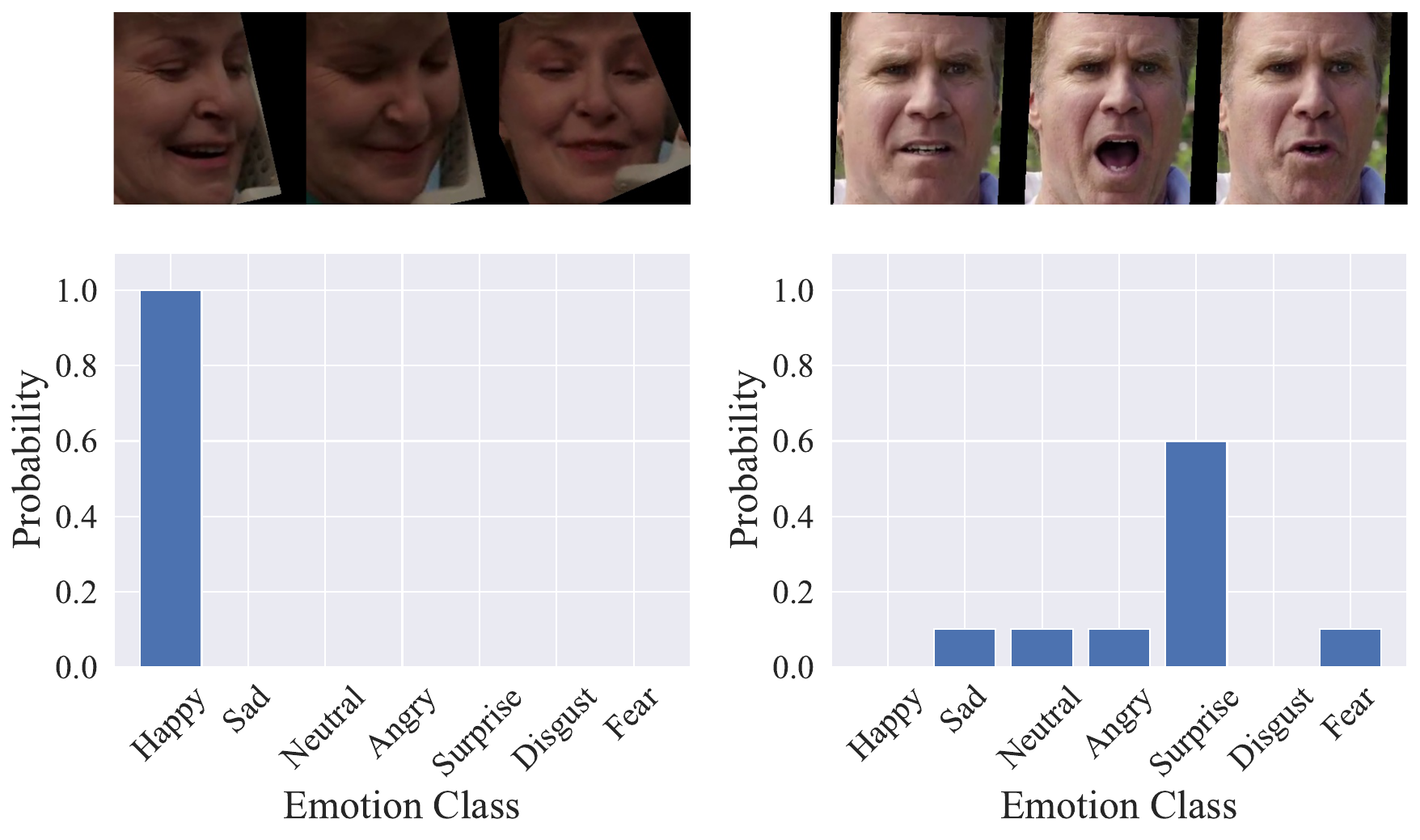}
  \vspace{0pt}
  \caption{Examples of a clear facial expression (left) and ambiguous facial expression (right) with their soft label annotations in the DFEW dataset~\protect\cite{jiang2020dfew}}
  \vspace{0pt}
  \label{fig:compare_dfew}
\end{figure}

Fig.~\ref{fig:compare_dfew} presents examples of facial expression images along with their corresponding emotion labels from the DFEW dataset. In the figure, the left and right panels illustrate examples of a clear expression that all annotators identified as ``Happy'' and an ambiguous facial expression, respectively. In the example of the ambiguous expression shown in the right panel, the votes from the annotators are divided among five emotion classes, although more than half of the annotators judged this sample as ``Sad.''\par

In Table~\ref{tab:ex-dist}, the distribution of emotion classes in the DFEW dataset is presented. The dataset is imbalanced, with a relatively higher number of samples labeled as ``neutral'' and ``happy,'' and fewer labeled as ``disgust'' and ``fear.''

Table~\ref{tab:voting_dist} displays the distribution of the maximum label vote count in DFEW, based on the highest number of votes received for each video clip. It should be noted that in the DFEW dataset, samples with a maximum label vote count of fewer than five were excluded by the authors.

\begin{figure}[t]
  \centering
  \includegraphics[width=0.95\linewidth]{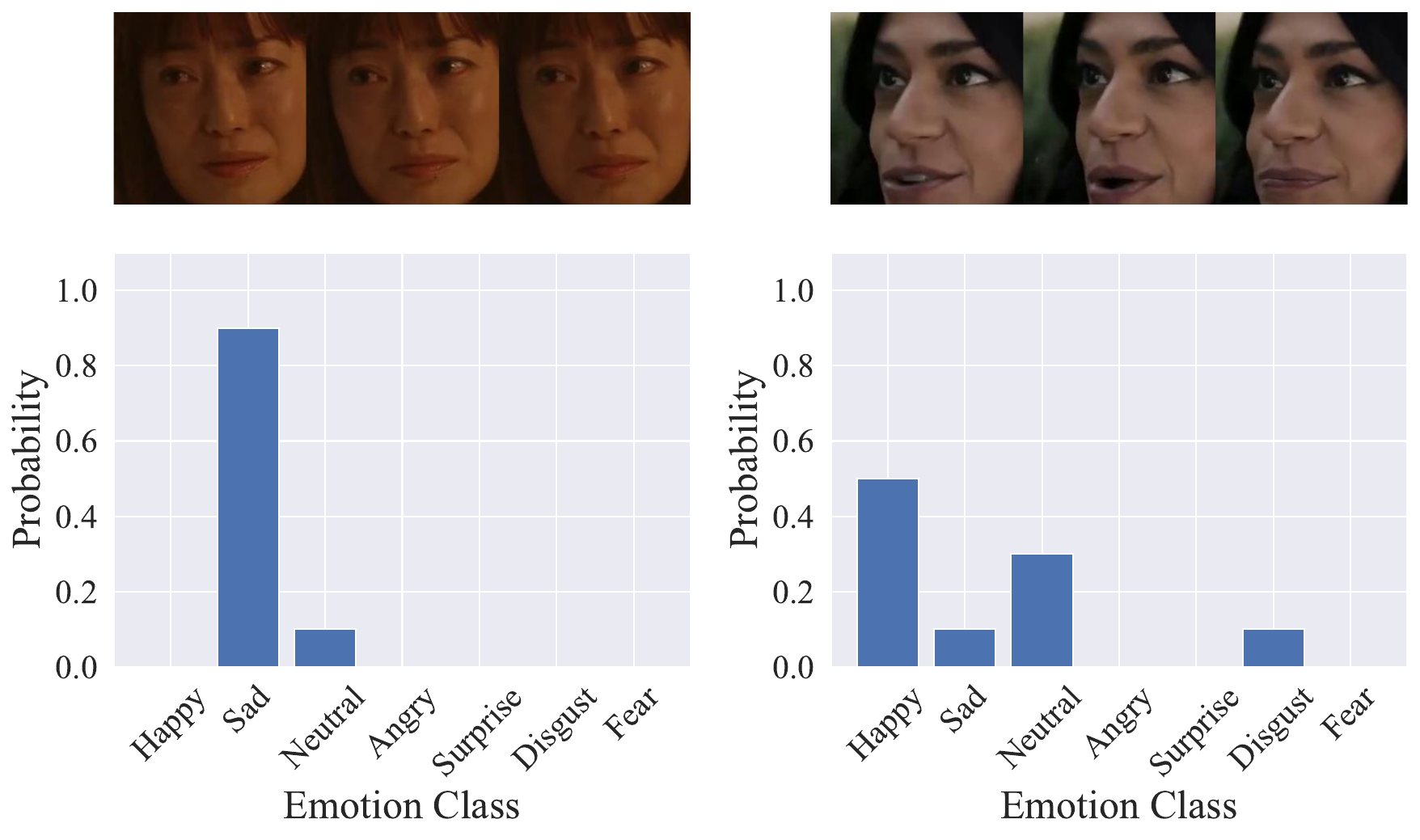}
  \vspace{0pt}
  \caption{Examples of a clear facial expression (left) and ambiguous facial expression (right) with their soft label annotations in the FERV39k-Plus dataset}
  \vspace{0pt}
  \label{fig:compare_ferv}
\end{figure}

{\color{mycolor}
\subsubsection{FERV39k-Plus}
We developed a new dataset, FERV39k-Plus, by re-annotating each video clip of the FERV39k dataset with soft labels of the seven emotion classes used in the DFEW dataset. This dataset will be made publicly available soon. 
FERV39k, detailed in Wang et al.~\cite{wang2022ferv39k}, is an in-the-wild dynamic facial expression recognition dataset. 
The video clips are categorized into four scenarios: Daily Life, Weak-Interactive Shows, Strong-Interactive Activities, and Anomaly Issues, which are further subdivided into 22 scenes with distinct characteristics. Similar to the DFEW dataset, FERV39k features a variety of challenging interferences that reflect practical scenarios. However, unlike the DFEW dataset, FERV39k was not annotated with soft labels for its video clips.

Examples from the FERV39k-Plus dataset are shown in Fig.~\ref{fig:compare_ferv}. To construct the FERV39k-Plus dataset, annotations were collected from multiple annotators recruited via a crowdsourcing platform, Amazon Mechanical Turk. The following approach was employed to ensure data quality and reliability: First, annotators who had at least 500 approved tasks in the past and an approval rate of over 98\% were recruited. As a result, 2,390 annotators participated. Second, each video clip was assigned to ten annotators, who were instructed to select a single emotion from seven predefined classes. 
Workers received the following instruction: ``Choose the correct category for the facial expression in the video. Descriptions for each category are provided on the left.''
The user interface follows the design used in~\cite{wang2022ferv39k}, with a modification in which only the seven emotion class names were provided as options, unlike in the original implementation.
No restrictions were placed on the number of video clips an annotator can evaluate, leading to variations in the number of annotations per worker. Some annotators labeled multiple clips, while others contributed to only a few, with the highest number of annotations being 1,584 and the lowest being one.
Finally, video clips without a single most-voted class were removed to ensure accurate metric calculations consistent with existing studies. For example, if 50\% of annotators select ``Happy'' and the remaining 50\% choose ``Surprise,'' the clip is excluded from the dataset.
The proportion of annotators who voted for each class was used to generate soft labels, while the class receiving the highest number of votes is assigned as the hard label.

The basic statistics of the FERV39k-Plus dataset are summarized as follows. The dataset contains a total of 33,467 video clips. Table~\ref{tab:ex-dist} shows the distribution of emotion classes in FERV39k-Plus, where each class corresponds to the emotion receiving the highest number of votes from annotators for each video. The most frequent emotion class is ``Neutral,'' followed by ``Happy'' and ``Sad,'' collectively accounting for approximately 80\% of the dataset. The least frequent class is ``Surprise,'' representing less than 1\% of the total video clips.
In addition, Table~\ref{tab:voting_dist} presents the distribution of the maximum label vote count in FERV39k-Plus. The distribution indicates that the most common maximum vote count is four, suggesting that annotators seldom reach unanimous agreement. This observation highlights the inherent ambiguity present in facial expression annotations.

}


\begin{table}[!t]
\caption{Emotion class distribution (\%) of the DFEW and FERV39k-Plus datasets}
\centering
    \label{tab:ex-dist}
    \vspace{0mm}
    \scalebox{1}[1]{
    {\tabcolsep=2pt
    
    \begin{tabular}{c|cccccccc} 
    \toprule
         dataset & Happy & Sad & Neutral & Angry & Surprise & Disgust & Fear \\ 
        \midrule
            DFEW  & 20.90 & 16.19 & 22.82 & 18.58 & 12.56 & 1.24 & 7.71   \\
            FERV39k-Plus   & 24.02 & 21.67 & 34.05 & 13.06 & 0.81 & 4.73 & 1.65  \\
    \bottomrule
    \end{tabular}}
    }
   \vspace{0pt}
\end{table}

\begin{table}[!t]
\caption{Distribution (\%) of the maximum vote count per video clip in the DFEW and FERV39k-Plus datasets. Video clips with a maximum vote count of less than two are not present in either dataset.}

\centering
    \label{tab:voting_dist}
    \vspace{0mm}
    \scalebox{1}[1]{
    {\tabcolsep=2pt
    
    \begin{tabular}{c|cccccccccc} 
    \toprule
         Max Votes &  3 & 4 & 5 & 6 & 7 & 8 & 9 & 10 \\ 
        \midrule
            DFEW  &  0.00 & 0.00 & 0.00 & 19.78 & 17.92 & 17.29 & 18.39 & 26.62   \\
            FERV39k-Plus   &  8.73 & 26.93 & 23.78 & 16.86 & 10.92 & 7.27 & 3.92 & 1.57  \\
    \bottomrule
    \end{tabular}}
    }
   \vspace{0pt}
\end{table}


\subsection{Experimental conditions}\label{sec:exp-condition}

{\color{mycolor}
\subsubsection{Evaluation Methodology}
In this experiment, we employ new Train-Validation-Test splits for both DFEW and FERV39k-Plus datasets to achieve a more practical evaluation scenario.
Previous studies have adopted a less effective method for evaluating their techniques on the DFEW and FERV39k datasets. These datasets typically come with predefined Train-Test splits, and prior research has frequently chosen the best models based on test split performance. However, a more effective evaluation of generalization involves selecting models based on their performance during training and then testing these models on the test split to assess their ability to handle unseen data, which can be unfair. Reviewing the released codes from these studies reveals that many selected their top-performing models by directly assessing test scores.

To better reflect practical applications and enable fair comparisons, this study revises the conventional Train-Test split strategy by introducing a Train-Validation split within the original training data. Specifically, we divide the original Train split into new Train and Validation subsets using an 8:2 ratio. This division is performed in a stratified manner to maintain the class distribution across both subsets in accordance with the original Train split. The Test split remains unchanged.

Model selection is based on performance evaluated on the Validation split. For the DFEW dataset, which includes predefined five-fold cross-validation splits, we apply the same Train-Validation division within the training split of each fold. This methodological adjustment, particularly in the handling of training data, distinguishes our approach from those used in prior studies.

\subsubsection{Architecture}
To assess the performance of MIDAS, we utilized the MAE-DFER model~\cite{sun2023mae}. The MAE-DFER is a masked autoencoder-type architecture that incorporates the Efficient Local-Global Interaction Transformer as its encoder. This model was pre-trained on the VoxCeleb2~\cite{chung2018voxceleb2} dataset. In our experiments, we adhered to the hyperparameter settings outlined in~\cite{sun2023mae}, with the exception of the batch size, which we adjusted to 64.
%
The input image of each frame was resized to 160 $\times$ 160. For the loss function, we used cross-entropy loss. In addition, $\alpha$ for the beta distribution for MIDAS was set to 0.8.\par

\subsubsection{Baseline methods}
Given our definition of new train-val-test splits for the DFEW dataset and the creation of the new dataset FERV39k-Plus, it is imperative to evaluate the performance of existing methods under these revised settings. In our experiments, we meticulously replicated the methods of earlier studies and evaluated them within our revised framework. 
We should note that some of the existing works are not publicly available and we re-implement available codes for our evaluation.
We implemented several comparison methods using available codes, including Former-DFER~\cite{zhao2021former}, GCA+IAL~\cite{li2023intensity}, Wang et al.~\cite{wang2023rethinking}, and S2D~\cite{chen2024S2D}. Additionally, we assessed various video recognition models, namely the Temporal Shifted Module (TSM) with a ResNet-18 backbone, and VideoMAE.
The TSM~\cite{lin2019tsm} is recognized for its hardware efficiency in recognizing facial emotions from video clips. The ResNet-18 backbone of this model was pre-trained on the ImageNet database\cite{russakovsky2015imagenet}. 
VideoMAE~\cite{tong2022videomae} extends the masked autoencoder concept to dynamic data and uses customized video tube masking at a very high ratio; for our experiment, we pre-trained this model on the VoxCeleb2 dataset. For a fair comparison, we excluded multi-modal methods from our evaluation.

\subsubsection{Evaluation indexes}
For the evaluation indexes, we employed the unweighted average recall (UAR) and weighted average recall (WAR), which are officially used in~\cite{jiang2020dfew,wang2022ferv39k}. UAR represents the average prediction accuracy of each class and WAR represents accuracy. For the evaluation of the DFEW dataset, we calculated the averages of UAR and WAR over five groups of cross-validation. 

\subsection{Result}\label{ambiguity}\label{sec:ex-res}
\subsubsection{Result on DFEW}
Table~\ref{tab:each_class_dfew} summarizes the accuracy for each emotion class, the WAR, and UAR.
The table compares the performance of models trained on the original video data without augmentation using hard labels and our proposed model trained with MIDAS.
Our model achieves the highest scores in both WAR and UAR. Regarding per-class accuracy, MIDAS outperforms all other methods across all emotion classes.
In particular, when compared to MAE-DFER under the hard-label condition, which marks the second highest UAR and WAR, our method significantly improves accuracy for ``Disgust'' (MIDAS: 20.00, Hard label:11.72) and ``fear''(MIDAS: 40.02, Hard label:36.14). These two classes have considerably fewer training samples than the others, as illustrated in Table~\ref {tab:ex-dist}.

{\renewcommand{\arraystretch}{1.2}
\begin{table*}[t]
\centering
    \caption{Comparison of the accuracy for each emotion class, UAR, and WAR on \textbf{DFEW} dataset. Bold and underlined scores denote the best and second best, respectively.}
    \vspace{0pt}
    \label{tab:each_class_dfew}
    \resizebox{\textwidth}{!}{%
    \scalebox{0.9}[0.9]{
    \begin{tabular}{p{0.3\linewidth}|c|ccccccc|c|c} 
    \toprule
         \multirow{2}{*}{Method} &\multirow{2}{*}{Label} &  \multicolumn{7}{c}{Accuracy for each emotion class (\%)} & \multicolumn{2}{|c}{Metrics} \\ \cmidrule(lr){3-9}\cmidrule(lr){10-11}
             & & Happy & Sad & Neutral & Angry & Surprise & Disgust & Fear & UAR & WAR\\
     \midrule  
         Former-DFER~\cite{zhao2021former}& Hard  & 83.31 & 58.29 & 66.13 & 66.96 & 52.35 & 0.00 & 26.39 & 50.44 & 62.99 \\
         
        GCA+IAL~\cite{li2023intensity} 
        & Hard  & 85.88 & 62.57 & 67.79 & 73.42 & 59.35 & 0.00 & 24.86 &  53.41 & 66.58\\
          Wang et al.~\cite{wang2023rethinking} & Hard  &  84.95 & 65.37 & 58.80 & 69.08 & 54.34 & \underline{16.81} & 34.40 &  53.06 & 63.94\\
          S2D~\cite{chen2024S2D}& Hard  & \underline{92.18} & \bf{79.98} & 73.25 & 76.00 & 57.36 & 2.08 & \underline{39.49} & 60.05 & \underline{73.34} \\
         TSM~\cite{lin2019tsm} & Hard  &  85.40& 59.40& 60.21& 67.79& 48.47& 2.07& 20.62& 49.14& 61.50 \\
         VideoMAE~\cite{tong2022videomae}& Hard &  87.48 & 71.91 & \underline{74.19} & 73.56 & \underline{62.74} & 15.17 & 36.70 &  58.50 & 68.97 \\

         MAE-DFER~\cite{sun2023mae}& Hard &  92.02 & 73.60 & 72.50 & \underline{77.61} & 61.44 & 11.72 & 36.14 & \underline{60.76} & 72.76 \\
         MAE-DFER~\cite{sun2023mae} (Ours) & MIDAS &  \bf{92.88} & 75.18 & \bf{76.85} & \bf{78.03} & \bf{62.81} & \bf{20.00} & \bf{40.02} &  \bf{63.68} & \bf{74.84}\\
    \bottomrule
    \end{tabular}
    }}
\end{table*}
}

\subsubsection{Result on FERV39k-Plus}
{\color{mycolor}
Table~\ref{tab:each_class_ferv} shows the accuracy for each emotion class, the WAR, and the UAR of baseline methods and our model trained with MIDAS on FERV39k-Plus dataset.
Our model with MIDAS achieves the highest UAR (45.28) and WAR (68.65), surpassing all other methods. It outperforms competing approaches in terms of accuracy for the ``Sad,''``Angry,'' and ``Fear'' classes. 

For the ``Surprise'' class, which has the fewest samples, the accuracy of our model (5.13) is lower than that achieved by Wang et al.~\cite{wang2023rethinking} (14.63), who obtained the highest score. However, the difference in the number of correctly classified samples is small, indicating that the variation in model accuracy is not substantial.

Compared to MAE-DFER with hard labels, our model outperformed in UAR. In addition, the accuracy of our model for each class is improved or the same except for ``Disgust'' class.
The reason for this lower accuracy in ``Disgust'' class is discussed in Sec.~\ref{sec:ab_coexist}.}
{\renewcommand{\arraystretch}{1.2}
\begin{table*}[t]
\centering
    \caption{Comparison of the accuracy for each emotion class, UAR, and WAR on \textbf{FERV39k-Plus} dataset. Bold and underlined scores denote the best and second best, respectively.}
    \vspace{0pt}
    \label{tab:each_class_ferv}
    \resizebox{\textwidth}{!}{%
    \scalebox{0.9}[0.9]{
    \begin{tabular}{p{0.3\linewidth}|c|ccccccc|c|c} 
    \toprule
         \multirow{2}{*}{Method} &\multirow{2}{*}{Label} &  \multicolumn{7}{c}{Accuracy for each emotion class (\%)} & \multicolumn{2}{|c}{Metrics} \\ \cmidrule(lr){3-9}\cmidrule(lr){10-11}
             & & Happy & Sad & Neutral & Angry & Surprise & Disgust & Fear & UAR & WAR\\
     \midrule  
         Former-DFER~\cite{zhao2021former}& Hard  & 75.19 & 68.08 & 71.57 & 42.33 & \underline{7.41} & 0.00 & 5.93 & 38.64 & 62.75 \\
         
        GCA+IAL~\cite{li2023intensity} 
        & Hard  & 75.41 & 69.52 & 72.77 & 33.83 & 0.00 & 0.00 & 0.00 &   35.93 & 62.21\\
          Wang et al.~\cite{wang2023rethinking} & Hard  &  \bf{84.00} & \underline{72.99} & 45.36 & 48.58 & \bf{14.63} & 0.00 & \underline{16.48} &  39.32 & 57.89\\
         S2D~\cite{chen2024S2D} & Hard & 80.93 & 72.96 & \bf{77.96} & \bf{50.08} & 2.56 & 0.00 & 8.70 & 41.88 & \underline{68.50}\\
         TSM~\cite{lin2019tsm} & Hard  &  75.86& 66.80& 75.50& 36.74& 0.00& \underline{3.52} & 1.69& 37.20&63.31 \\
         VideoMAE~\cite{tong2022videomae}& Hard & 77.35 & 69.27 & 77.38 & 38.88 & 0.00 & 0.00 & 1.09 &   37.71 & 65.21 \\
         MAE-DFER~\cite{sun2023mae}& Hard &  78.87 & 68.09 & 70.59 & \underline{49.02} & 5.13 & \bf{5.76} & 13.04 &  \underline{43.28} & \bf{68.65} \\
         MAE-DFER~\cite{sun2023mae} (Ours) & MIDAS &  \underline{81.24} & \bf{73.64} & \underline{77.43} & \bf{50.08} & 5.13 & 0.36 & \bf{20.65} &   \bf{45.28} & \bf{68.65}\\
    \bottomrule
    \end{tabular}
    }}
\end{table*}
}

%% file: src/5_AblationStudy.tex
\label{sec:ablation}
{\color{mycolor}
\subsection{Comparison of our method with different architectures}
{\renewcommand{\arraystretch}{1}
\begin{table}[t]
\caption{Comparison of our method with different architectures}
\centering
    \label{tab:ab_diff_arch}
    \vspace{0mm}
    \scalebox{1}[1]{
    \begin{tabular}{c|c|cc|cc} 
    \toprule
         \multirow{1}{*}{Method} &\multirow{1}{*}{Label} &  \multicolumn{2}{c}{DFEW} & \multicolumn{2}{|c}{FERV39k-Plus} \\ \cmidrule(lr){3-6}
             & & UAR & WAR & UAR & WAR \\
        \midrule
               & Hard &  49.14 & 61.50  & 37.20 & 63.31\\
            TSM~\cite{lin2019tsm}   & Soft &  51.64&64.71 & 38.60 &63.43\\
               & MIDAS & 52.84 & 65.26 & 39.00 & 64.20 \\
        \midrule
               & Hard &  58.50 & 68.97 & 37.71 & 65.12\\
            VideoMAE~\cite{tong2022videomae}   & Soft & 59.51 & 70.86 & 39.14 & 66.26\\
               & MIDAS & 60.25 & 71.42 & 43.65 & 67.49 \\
        \midrule
                       & Hard & 60.76 & 72.76 & 43.15 & 68.65\\
            MAE-DFER~\cite{sun2023mae}   & Soft & 62.64 & 74.52 & 41.25 & 68.08\\
                       & MIDAS & 63.68 & 74.84 & 45.28 & 68.65 \\
        \bottomrule
        \end{tabular}
    }
   \vspace{0pt}
\end{table}
}
We conducted an experiment using different architectures for DFER to evaluate the effectiveness of our proposed method. In the experiment in Sec.~\ref{sec:experiments}, we applied our method to MAE-DFER~\cite{sun2023mae}, a model designed for dynamic facial expression recognition, and observed performance improvements. However, the applicability of our method to other deep learning architectures was not examined. To address this, we compared the performance of TSM, MAE, and MAE-DFER with and without our method.

Additionally, to assess the effectiveness of our approach beyond training with soft labels alone, we compared these architectures when trained with soft labels to those incorporating MIDAS. All models were trained using the same experimental settings as in previous experiments, ensuring consistency in evaluation. 

The results of the experiment are shown in Table~\ref{tab:ab_diff_arch}. As indicated in the table, models trained with MIDAS achieve higher UAR and WAR compared to those trained with either hard labels or soft labels, irrespective of the architecture used.

}
{}

\subsection{Our method with hard labels}\label{sec:ab-hard}
We investigated whether applying a mixing strategy to dynamic facial expression data annotated with hard labels could enhance model performance. Although our proposed method improves recognition accuracy, the process of generating soft labels requires considerable resources. Therefore, if training with single-label annotations, which are easier to obtain, also yields performance improvements, it would provide a more practical alternative.

The experimental procedure involved applying our method directly to video clips annotated with hard labels instead of soft labels. The modified dataset was then used to train the MAE-DFER model, with the combined labels normalized according to the training settings described previously.

The results, presented in Table~\ref{tab:ab-midas-with-hard}, show that the model trained with MIDAS on hard labels improved UAR by 0.65\% on the DFEW dataset and by 0.13\% on the FERV39k-Plus dataset compared to the model trained solely on hard labels. For WAR, the model trained with MIDAS on hard labels achieved an improvement of 0.78\% on the DFEW dataset. However, on the FERV39k-Plus dataset, its WAR score (67.86) was lower than that of the model trained with hard labels alone (68.65).

These findings suggest that our strategy can enhance DFER performance when using hard labels, although certain limitations remain.

\begin{table}[t]
\caption{Comparison of the results of our method with and without hard label}
\centering
    \label{tab:ab-midas-with-hard}
    \vspace{0mm}
    \scalebox{1}[1]{
    \begin{tabular}{c|cc|cc} 
    \toprule
         \multirow{1}{*}{Label} &  \multicolumn{2}{c}{DFEW} & \multicolumn{2}{|c}{FERV39k-Plus} \\ \cmidrule(lr){2-5}
              & UAR & WAR & UAR & WAR \\
        \midrule
              Hard & 60.76 & 72.76 & 43.15 & 68.65\\
             MIDAS+Hard & 61.41 & 73.54 & 43.28 & 67.86\\
              MIDAS & 63.68 & 74.84 & 45.28 & 68.65 \\
        \bottomrule
        \end{tabular}
    }
   \vspace{0pt}
\end{table}
            

{\color{mycolor}
\subsection{Cross-dataset evaluation}
Generalization ability is critical in facial expression recognition, as it must account for variations in lighting conditions, facial orientations, and facial shapes. These factors can significantly impact the accuracy of recognition models in real-world scenarios.

To assess the generalization capability of our approach, we conducted a cross-dataset evaluation. In this experiment, models were trained on the DFEW dataset and tested on the FERV39k-Plus dataset, and vice versa, to evaluate their performance under first-time encountered conditions. We trained the TSM, VideoMAE, and MAE-DFER models using MIDAS, and compared their performance with those trained using only hard labels, as well as with existing architectures.
Importantly, these models were not further fine-tuned on either the DFEW or FERV39k-Plus datasets. Instead, they were directly evaluated on the test set of the other dataset to measure their out-of-sample generalizability.

Table~\ref{tab:cross} presents the UAR and WAR for each method on the FERV39k-S and DFEW test sets. In the results of the models trained on DFEW and tested on FERV39k-Plus, MIDAS exhibited superior performance in especially WAR metrics when compared to models trained solely on hard labels. 
As for the results of the models trained on FERV39k-Plus and tested on DFEW, MIDAS exhibited superior performance in both UAR and WAR metrics when compared to models trained solely on hard labels. 
This suggests that MIDAS may enhance the generalization capability of models to unseen conditions. 
Furthermore, the performance of the MAE-DFER when integrated with MIDAS surpassed that of the existing methods, including Former-DFER, GCA+IAL, and Wang et al~\cite{wang2023rethinking}.
{\renewcommand{\arraystretch}{1.2}
\begin{table}[t]
\caption{Cross-dataset evaluation between the DFEW and FERV39k-Plus datasets. D and F denote the DFEW and FERV39k-Plus datasets, respectively. D$\rightarrow$F indicates models trained on DFEW and tested on FERV39k-Plus, while F$\rightarrow$D refers to models trained on FERV39k-Plus and evaluated on DFEW.}

\centering
    \label{tab:cross}
    \vspace{0mm}
    \scalebox{1}[1]{
    \begin{tabular}{l|cc|cc} 
    \toprule
         \multirow{1}{*}{Method} &  \multicolumn{2}{c}{D $\rightarrow$ F} & \multicolumn{2}{|c}{F $\rightarrow$ D} \\ \cmidrule(lr){2-5}
             & UAR & WAR & UAR & WAR \\
        \midrule
            Former-DFER   & 35.27 & 44.23 & 33.31 & 44.72\\
            GCA+IAL & 31.89 & 42.24 & 32.50 & 43.82 \\
            TFace & 31.02 & 40.72 & 34.18 & 43.45 \\
            S2R & 40.88 & 52.56 & 41.51 & 54.39 \\
        \midrule
            TSM   &  31.18 & 37.61  & 32.70 & 43.9 \\
            TSM+MIDAS &  33.44 & 41.73& 35.00 & 46.02 \\
        \midrule
            MAE   & 35.98 & 46.53 & 35.72 & 47.66\\
            MAE+MIDAS & 35.51 & 48.62 & 39.39 & 51.35 \\
        \midrule
            MAE-DFER   & 39.72 & 52.98 & 39.87 & 52.05\\
            MAE-DFER+MIDAS & 39.59 & 54.82 & 42.23 & 54.60 \\
        \bottomrule
        \end{tabular}
    }
   \vspace{0pt}
\end{table}
}
}
\subsection{Analysis of the impact of coexisting emotion}\label{sec:ab_coexist}
We analyzed how the presence of coexisting emotions affects the performance in DFER. In the DFEW and FERV39k-Plus datasets, soft labels were generated based on votes from ten annotators. Since annotators do not always agree, some assign votes to emotion classes different from the one with the highest count. For instance, in Fig.~\ref{fig:ex}, while ``Sad'' received the most votes with six votes, ``Happy,'' ``Angry,'' ``Disgust,'' and ``Fear'' also received one vote. To understand how frequently co-occurring emotions influence model performance, we analyzed their effects on recognition accuracy.

\subsubsection{Analysis on DFEW dataset}
Fig.~\ref{fig:coex_dfew} presents the average ratio of coexisting emotions for each emotion class in DFEW dataset. For example, in the case of the leftmost emotion class ``Happy,'' a considerable number of annotators also voted for ``Neutral'' in instances where ``Happy'' received the highest number of votes. The values shown in the figure were obtained by averaging the soft label values of the samples belonging to each respective emotion class. Based on this figure, the following observations are made:
\begin{itemize}
\setlength{\parskip}{0cm}
\setlength{\itemsep}{0cm}
    \item ``Neutral'' frequently coexists with all other emotion classes.
    \item ``Happy'' has a low tendency to coexist with other emotion classes.
    \item ``Angry'' and ``disgust'' often appear together.
    \item ``Sad,'' ``surprise,'' and ``fear' frequently co-occur.
\end{itemize}

Fig.~\ref{fig:conf_dfew} displays the confusion matrix of the MAE-DFER model trained with MIDAS, illustrating the influence of emotion class coexistence on classification performance, as highlighted in Fig.~\ref{fig:coex_dfew}.  Since ``Neutral'' tends to co-occur with other emotion classes more frequently than any other classes, the model often misclassifies other classes as ``Neutral.'' Notably, a significant proportion of samples labeled as ``Disgust'' are incorrectly predicted as ``Neutral.'' One the other hand, ``Happy'' exhibits minimal coexistence with other emotions, and the model correctly classifies most instances belonging to this category, achieving an accuracy of 92.88\%. 

{\color{mycolor}
\subsubsection{Analysis on FERV39k-Plus dataset}

\begin{figure*}[!t]
\centering
    \begin{tabular}{cc}
        \begin{minipage}[b]{0.45\linewidth}
        \centering
        \includegraphics[width=0.99\linewidth]{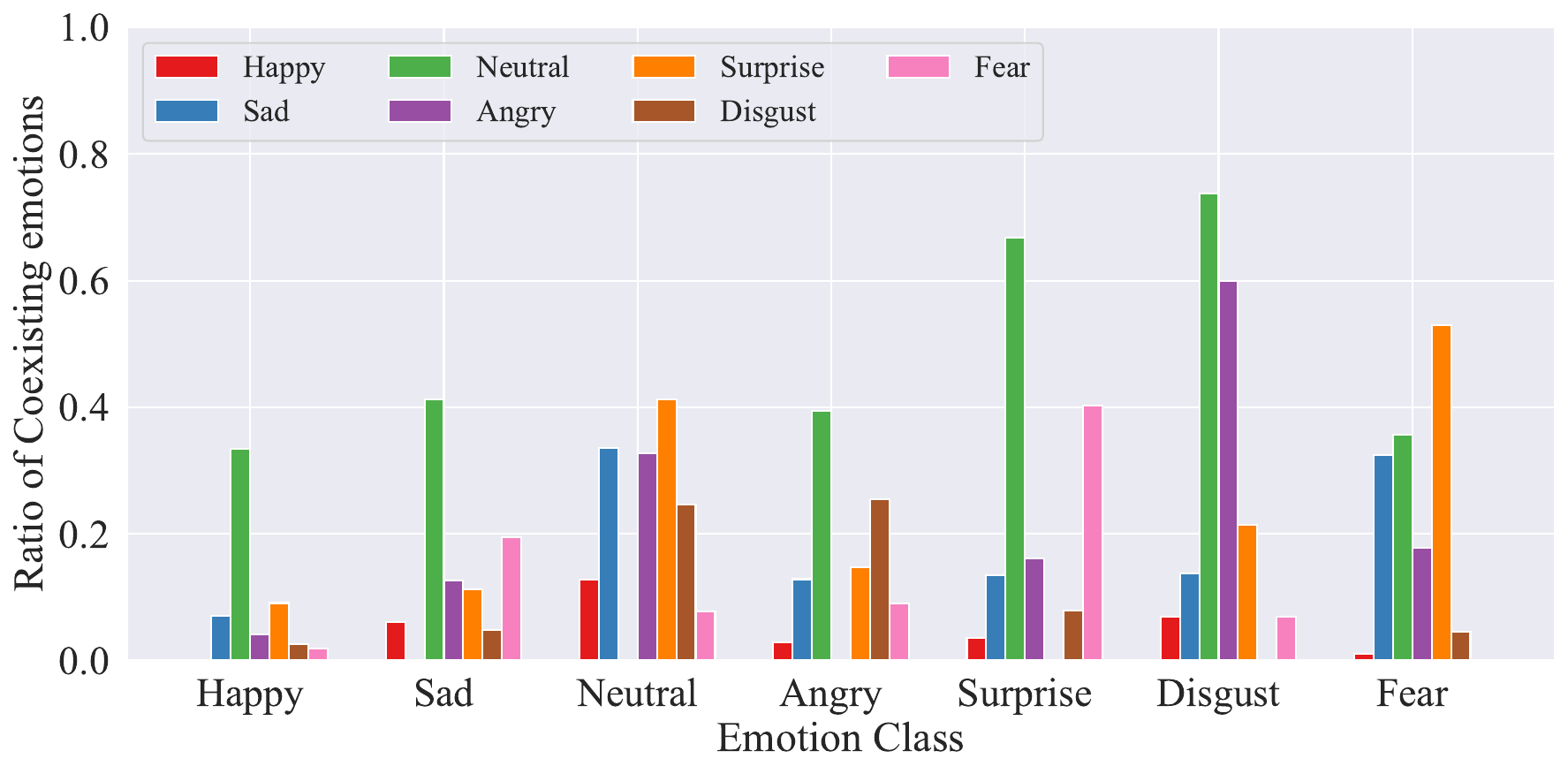}%
        \subcaption{Ratio of coexisting emotions for each emotion class in DFEW dataset~\cite{jiang2020dfew}}\label{fig:coex_dfew}%
        \end{minipage}%
        \hspace{10pt}
        \begin{minipage}[b]{0.45\linewidth}
        \centering
        \includegraphics[width=0.99\linewidth]{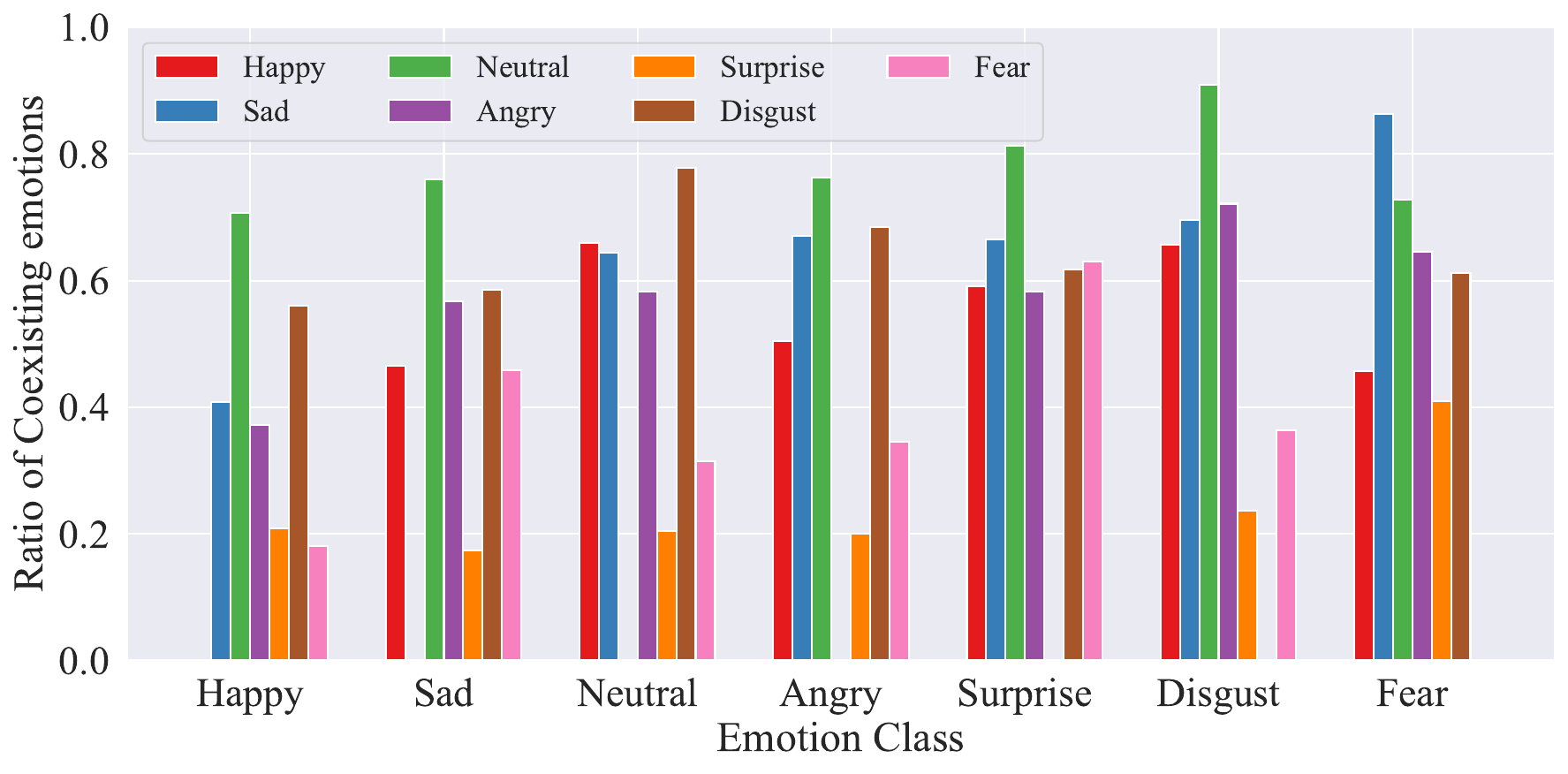}%
        \subcaption{Ratio of coexisting emotions for each emotion class in FERV39k-Plus dataset}\label{fig:coex_ferv}%
        \end{minipage}%
    \end{tabular}
    \caption{The ratio of coexisting emotions for each emotion class in the DFEW dataset~\cite{jiang2020dfew} and the FERV39k-Plus dataset is analyzed. The values in the figure were obtained by counting the occurrence of votes for each emotion class among the samples assigned to the corresponding emotion class. A higher value indicates a greater likelihood that annotators simultaneously voted for that emotion, suggesting a stronger tendency for it to coexist with other emotions.}
    \label{fig:multi}
    \vspace{-5pt}
\end{figure*}

\begin{figure*}[h!]
\centering
  \begin{tabular}{cc}
    \begin {minipage}{0.4 \textwidth}
      \centering
        \includegraphics [width=0.8\columnwidth] {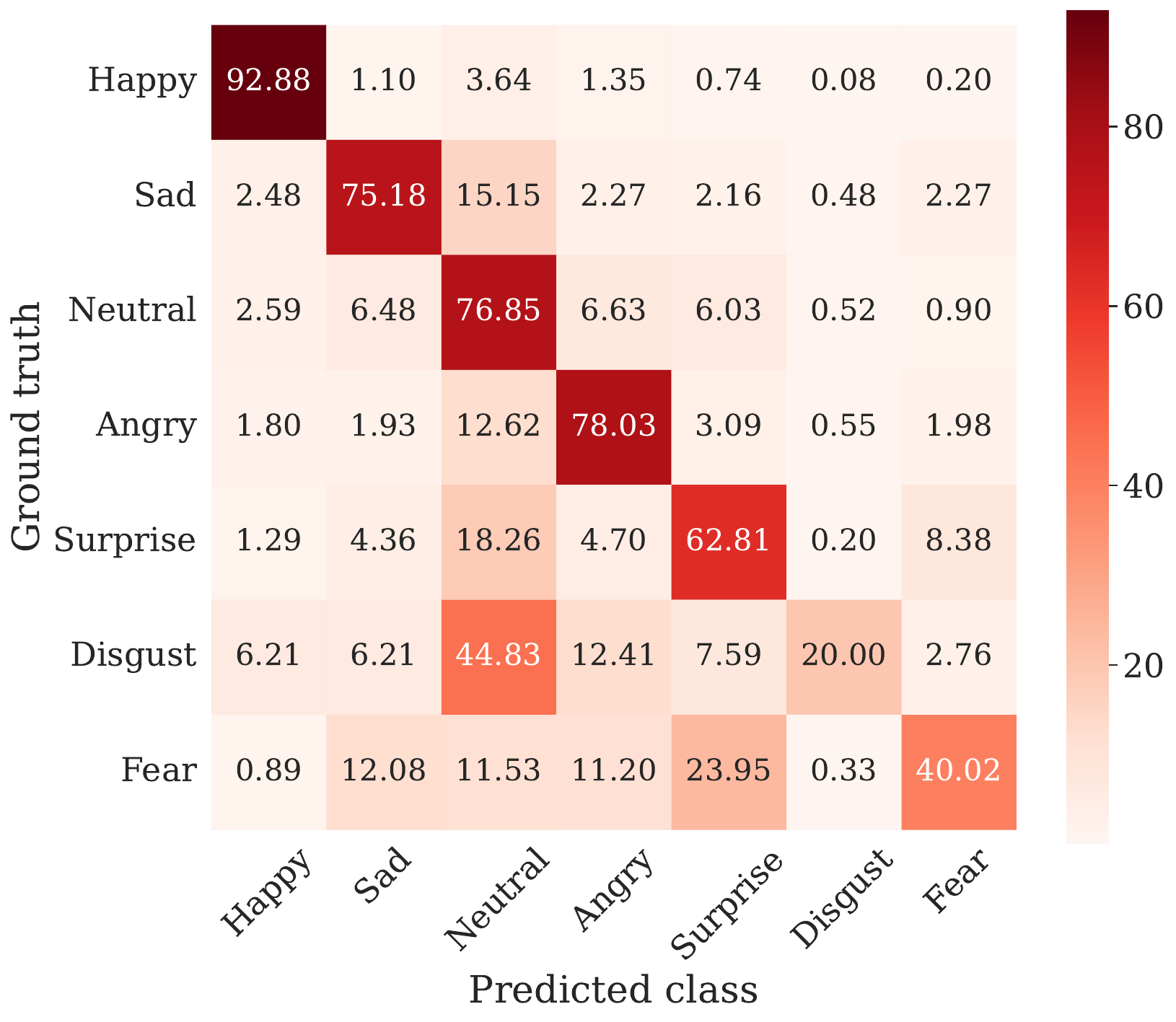}
        \subcaption{Confusion matrix on DFEW dataset~\cite{jiang2020dfew} dataset}\label{fig:conf_dfew}%
    \end {minipage}
    \begin{minipage}{0.4 \textwidth}
      \centering
        \includegraphics [width=0.8\columnwidth] {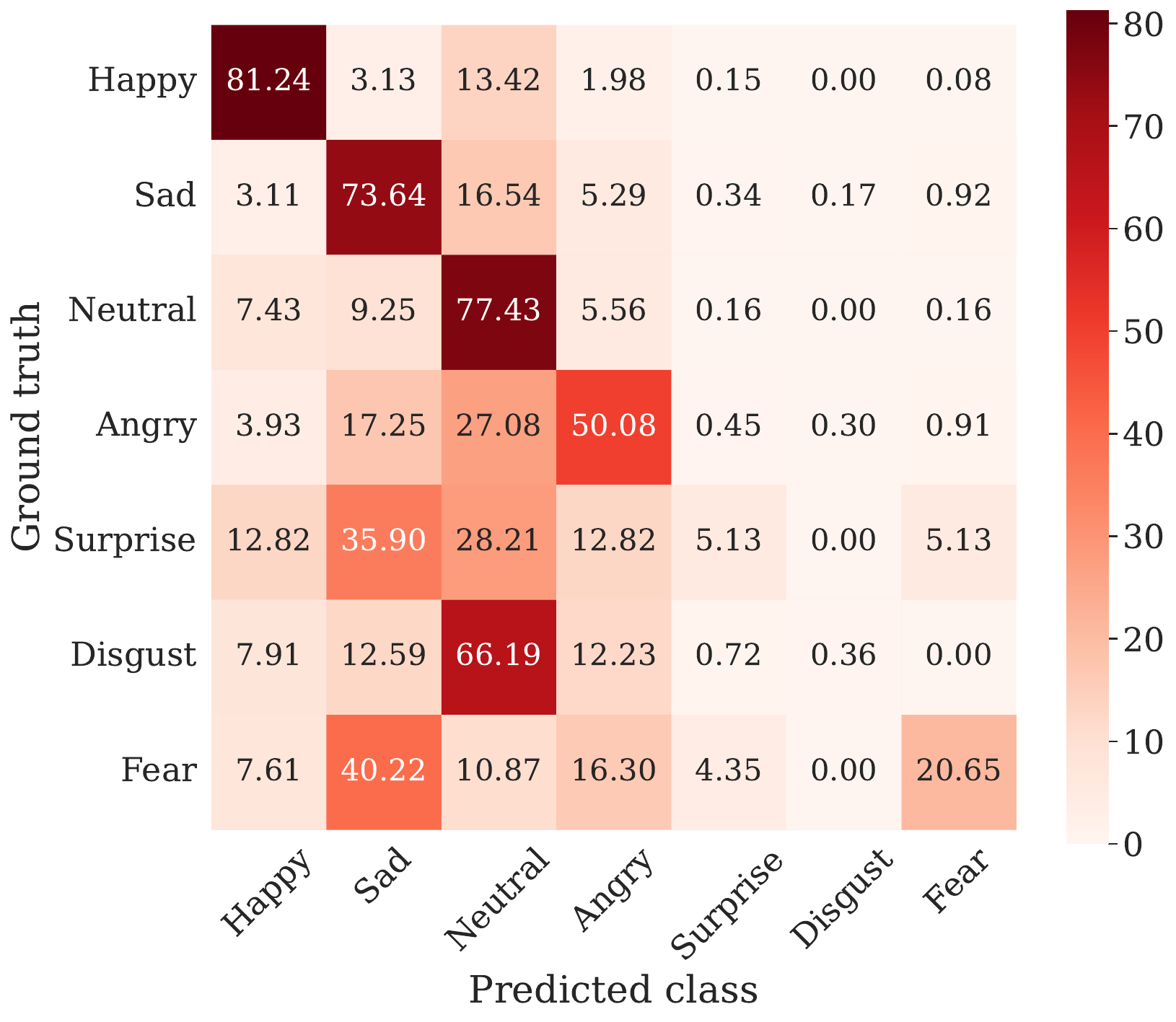}
        \subcaption{Confusion matrix on FERV39k-Plus dataset}\label{fig:conf_ferv}%
    \end {minipage}
  \end {tabular}
  \caption{Confusion matrix of MAE-DFER with MIDAS trained on DFEW dataset~\cite{jiang2020dfew} and FERV39k-Plus datasets}
\end {figure*}

\begin{figure*}[t]
  \centering
  \includegraphics[width=0.85\linewidth]{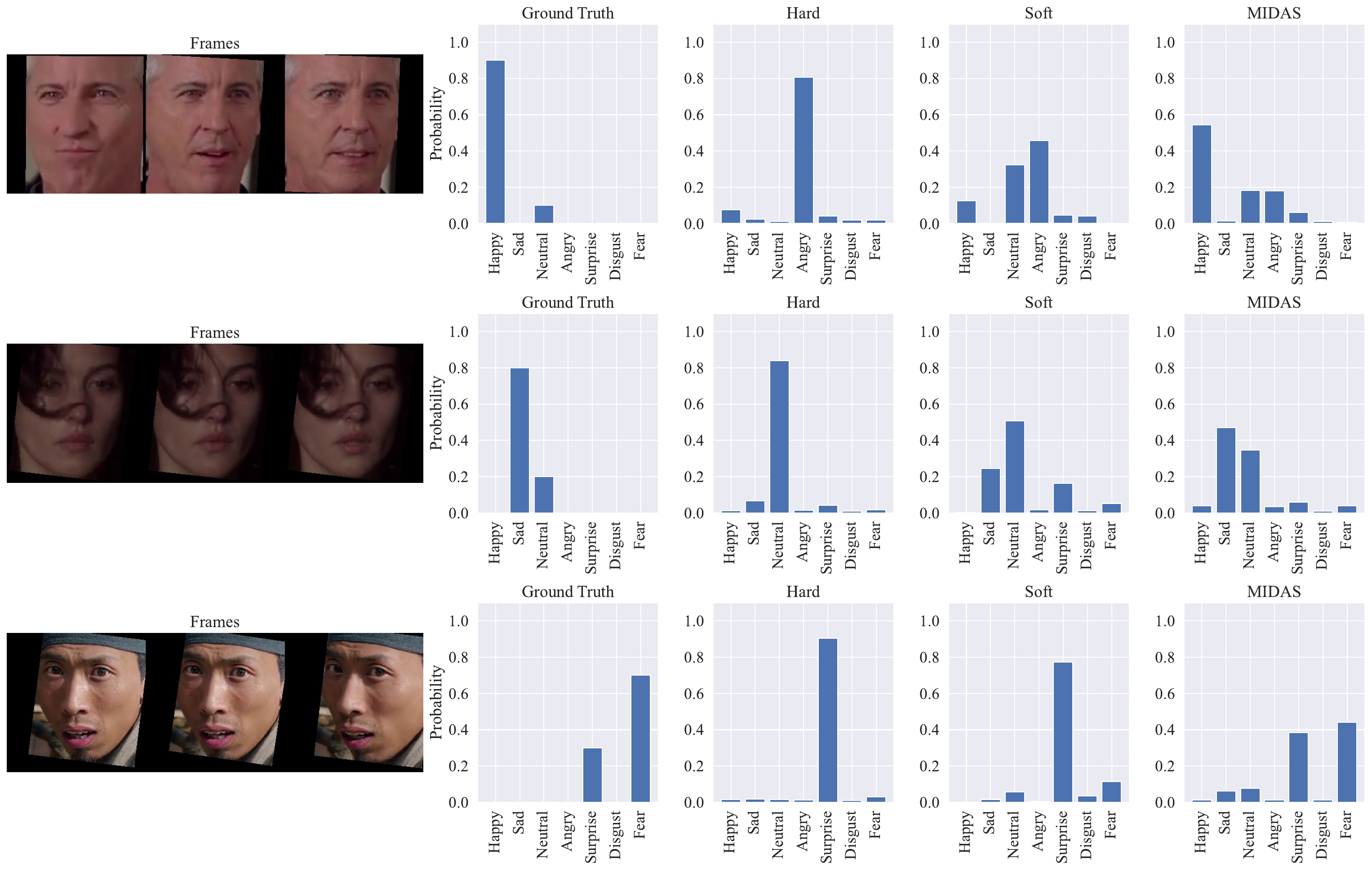}
  \vspace{0pt}
  \caption{Posterior probabilities and ground truth soft label on DFEW dataset~\cite{jiang2020dfew}.}
  \vspace{0pt}
  \label{fig:vis_result_dfew}
\end{figure*}
\begin{figure*}[!t]
  \centering
  \includegraphics[width=0.85\linewidth]{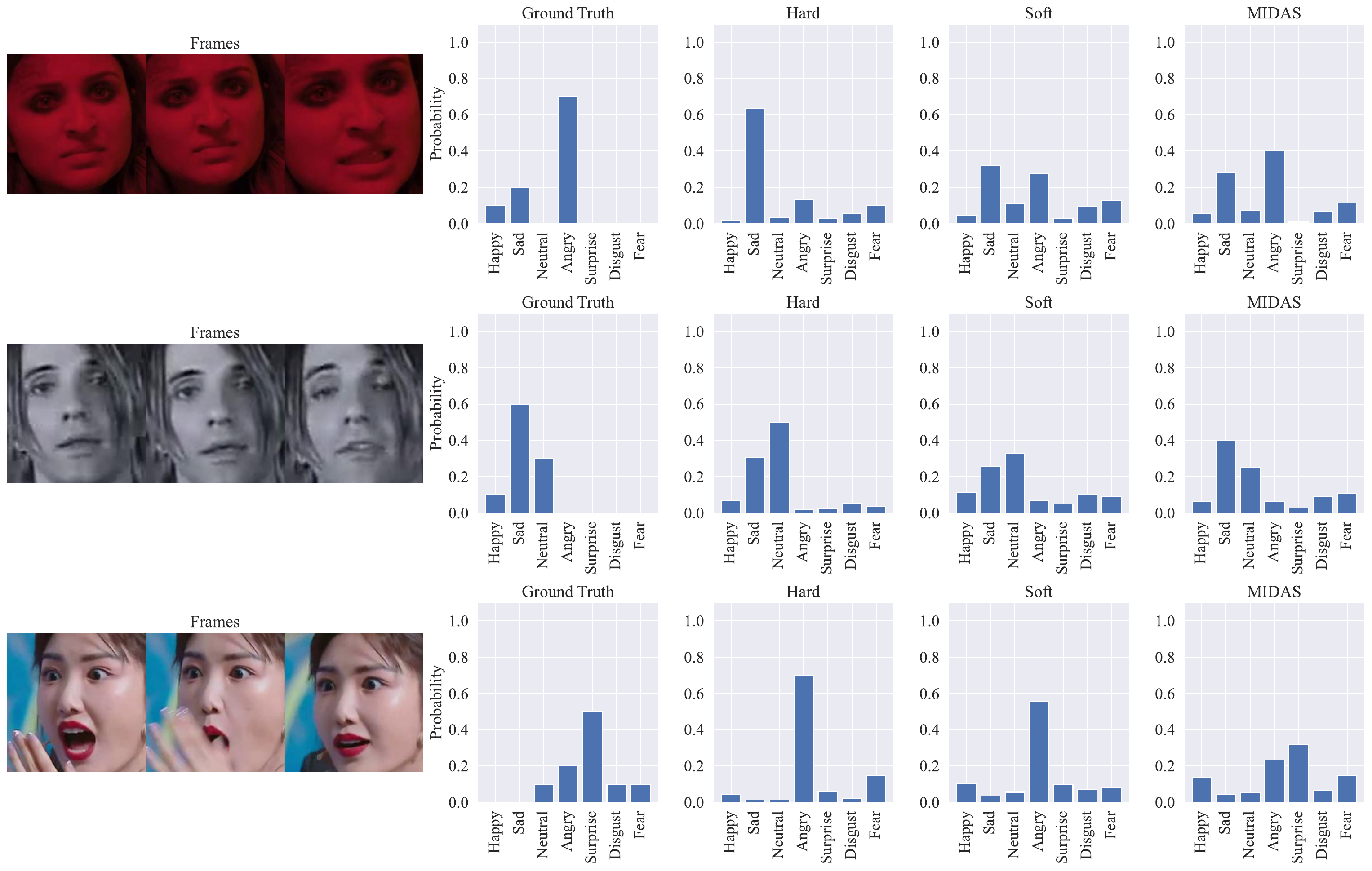}
  \vspace{0pt}
  \caption{Posterior probabilities and ground truth soft label on FERV39k-Plus dataset.}
  \vspace{0pt}
  \label{fig:vis_result_ferv}
\end{figure*}

Fig.~\ref{fig:coex_ferv} shows the average ratio of coexisting emotions for each emotion class of FERV39k-Plus dataset.  From this figure, the following are observed.
\begin{itemize}
\setlength{\parskip}{0cm}
\setlength{\itemsep}{0cm}
    \item ``Neutral'' and ``Sad'' tend to coexist with all other emotion classes.
    \item ``Neutral'' frequently coexists with ``Disgust,'' and vice versa.
    \item ``Sad'' often appears alongside ``Fear.''
\end{itemize}

Fig.~\ref{fig:conf_ferv} shows the confusion matrix of the MAE-DFER model trained with MIDAS on the FERV39k-Plus dataset, illustrating the impact of emotion class coexistence observed in Figure~\ref{fig:coex_ferv} on classification performance. Since ``Neutral'' and ``Sad'' frequently co-occur with other emotion classes, the model tends to misclassify samples from various classes as ``Neutral'' or ``Sad.'' Notably, instances labeled as ``Disgust'' are most often misclassified as ``Neutral,'' while those labeled as ``Fear'' are frequently recognized as ``Sad.'' 

}
\subsection{The effect of ambiguous data on model performance}
To confirm whether ambiguous data in the DFEW dataset affects the model performance, we investigated the effect of ambiguous data by comparing models trained on datasets with and without ambiguous data. In this comparison, we divided the original DFEW dataset into two groups: clear expression and mixed expression groups. The clear expression group consists of data with maximum soft label values of more than 0.9, e.g., the left example in Fig.~\ref{fig:compare_dfew}. The mixed expression group contains data regardless of the soft labels' values and includes ambiguous facial expressions such as the right example in Fig.~\ref{fig:compare_dfew}. To address the difference in data distribution, the distribution of each emotion class was matched to the original dataset by oversampling and down-sampling. Finally, the sizes of both data groups were set to an equal number (4275). We trained two models on these datasets with soft labels and evaluated them using a validation split of the original dataset. 

The results are shown in Table~\ref{tab:ab1}. The model with the mixed expression group obtained higher UAR and WAR scores than the model trained with the clear expression group. These results demonstrated that the existence of ambiguous data can improve the performance of DFER.

{\renewcommand{\arraystretch}{1.2}
\begin{table}[!t]
\centering
    \caption{Comparison of the results of models trained with and without ambiguous data. ``Mixed'' stands for the mixed expression group and ``Clear'' stands for the clear expression group.}
    \label{tab:ab1}
    \vspace{0mm}
    \scalebox{1}[1]{
    \begin{tabular}{c|c|c|c} 
    \toprule
         Method &Data & UAR & WAR \\
         \midrule
          \multirow{2}{*}{MAE-DFER}&Clear  & 56.06 & 69.82 \\
         
          &Mixed  & {58.67}& {71.38} \\
        \midrule
          \multirow{2}{*}{MAE-DFER+MIDAS}&Clear  & 56.63 & 70.53 \\
         
          &Mixed  & {60.86}& {73.01} \\
    \bottomrule
    \end{tabular}
    }
    \vspace{0pt}
\end{table}
} 

\begin{table}[!t]
\caption{Comparison of different $\alpha$ value}
\centering
    \label{tab:ab-alpha}
    \vspace{-2mm}
    \scalebox{0.8}[0.8]{
    \begin{tabular}{c|cccccccccc} 
    \toprule
         $\alpha$ & 0.1 & 0.2 & 0.3 & 0.4 & 0.5 & 0.6 & 0.7 & 0.8 & 0.9\\ 
        \midrule
            UAR  & 63.78 & 63.65 & 64.21 & 64.20 & 63.72 & 63.68 & 62.64 & 63.68 & 63.25  \\
            WAR   & 75.05 & 74.75 & 74.79 & 74.89 & 74.80 & 74.50 & 74.59 & 74.84 & 74.51  \\
    \bottomrule
    \end{tabular}
    }
   \vspace{0pt}
\end{table}

\subsection{Evaluation of alpha value}
The $\alpha$ value in the beta distribution plays a pivotal role in the mixing strategy~\cite{zhang2018mixup}.
To evaluate the impact of various $\alpha$ values on MIDAS, we carried out an experiment with $\alpha$ values selected from $\{0.1, 0.2, 0.3, 0.4, 0.5, 0.6, 0.7, 0.8, 0.9\}$. 
We used MAE-DFER as the backbone architecture for this experiment with the same settings for the remaining parameters.   
The results in Table~\ref{tab:ab-alpha} indicate that $\alpha = 0.4$ achieves the highest WAR, whereas $\alpha = 0.3$ yields the best UAR in MIDAS.

{\color{mycolor}
\subsection{Qualitative Evaluation}

We evaluate how well the model's output aligns with the distribution of ground truth soft labels. 
Fig.~\ref{fig:vis_result_dfew} and~\ref{fig:vis_result_ferv} illustrate the posterior probability distributions of MAE-DFER trained with hard labels, soft labels, and MIDAS for samples from the DFEW dataset~\cite{jiang2020dfew} and the FERV39k-Plus dataset. The input consists of image sequences.

For instance, in the second row of Fig.~\ref{fig:vis_result_dfew}, the posterior probability of the model trained with MIDAS assigns the highest likelihood to the ``Sad'' class, followed by ``Neutral.'' This prediction closely matches the ground truth soft label for the corresponding sample. A similar pattern is observed in the posterior probability distributions and sample classifications shown in Fig.~\ref{fig:vis_result_ferv}.
These results suggest that MIDAS improves the model's ability to predict emotion distributions more accurately.
}

%% file: src/6_Conclusion.tex
\vspace{0pt}
In this study, we proposed MIDAS, a data augmentation method designed to handle ambiguous facial expressions in DFER. MIDAS utilizes data mixing with soft labels by combining two video clips of facial expressions and their corresponding soft labels, generating diverse emotion combinations with varying intensities. To assess the effectiveness of our approach, we constructed a new dataset by assigning soft labels to an existing dataset and conducted experiments using both this newly created dataset and a publicly available DFER dataset. The experimental results demonstrated that MIDAS improves DFER performance, surpassing the state-of-the-art method. Additionally, the findings confirmed the effectiveness of MIDAS across different model architectures.
In future work, we plan to evaluate MIDAS using other domains of datasets, as it was evaluated using only the DFEW dataset. Although the MIDAS is developed for DFER with ambiguous facial expressions, it would be effective for other tasks that involve ambiguous class categorization with soft-labeled annotations.

